\documentclass[10pt,twocolumn,letterpaper]{article}

\usepackage{cvpr}              
\usepackage{mathrsfs}
\usepackage{svg}
\usepackage{stfloats}
\usepackage{multicol}

\definecolor{cvprblue}{rgb}{0.21,0.49,0.74}
\usepackage[pagebackref,breaklinks,colorlinks,allcolors=cvprblue]{hyperref}

\def\paperID{25} 
\def\confName{CVPR}
\def\confYear{2025}

\title{MFSR-GAN: Multi-Frame Super-Resolution with Handheld Motion Modeling}

\author{Fadeel Sher Khan\thanks{Work completed during an internship at Samsung Research America}\\
The University of Texas at Austin\\
{\tt\small fadeelkhan@utexas.edu}
\and
Joshua Ebenezer\\
Samsung Research America\\
{\tt\small j.ebenezer@samsung.com}
\and
Hamid Sheikh\\
Samsung Research America\\
{\tt\small hr.sheikh@samsung.com}
\and
Seok-Jun Lee\\
Samsung Research America\\
{\tt\small seokjun1.lee@samsung.com}
}

\begin{document}
\maketitle
\begin{abstract}
Smartphone cameras have become ubiquitous imaging tools, yet their small sensors and compact optics often limit spatial resolution and introduce distortions. Combining information from multiple low-resolution (LR) frames to produce a high-resolution (HR) image has been explored to overcome the inherent limitations of smartphone cameras. Despite the promise of multi-frame super-resolution (MFSR), current approaches are hindered by datasets that fail to capture the characteristic noise and motion patterns found in real-world handheld burst images. In this work, we address this gap by introducing a novel synthetic data engine that uses multi-exposure static images to synthesize LR-HR training pairs while preserving sensor-specific noise characteristics and image motion found during handheld burst photography. We also propose MFSR-GAN: a multi-scale RAW-to-RGB network for MFSR. Compared to prior approaches, MFSR-GAN emphasizes a “base frame” throughout its architecture to mitigate artifacts. Experimental results on both synthetic and real data demonstrates that MFSR-GAN trained with our synthetic engine yields sharper, more realistic reconstructions than existing methods for real-world MFSR.
\end{abstract}
\section{Introduction}
\label{sec:intro}
\begin{figure}[h]
    \setlength{\belowcaptionskip}{-10pt}
    \setlength{\abovecaptionskip}{-5pt}
    \centering
    \includegraphics[width=\linewidth]{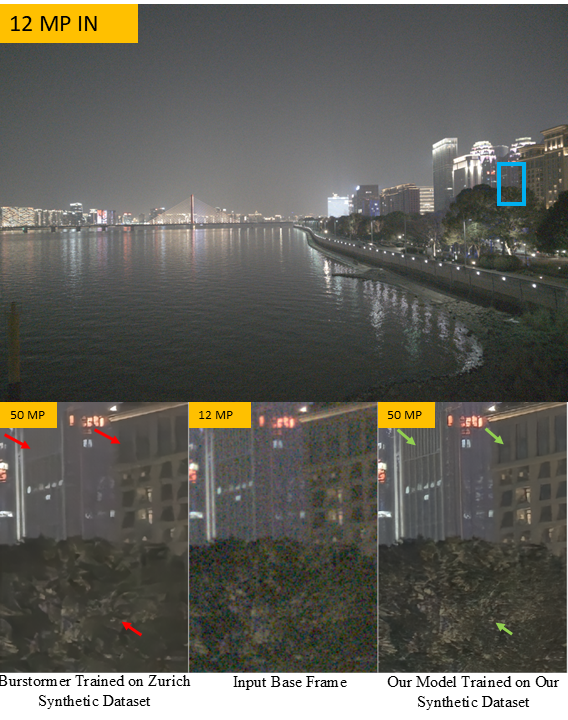}
    \label{fig:mpinet_overview}
    \caption{We train a novel multi-frame RAW to RGB super-resolution network on a novel synthetic data engine that generates sharper and more detailed images than the existing state-of-the-art model \cite{dudhane_burstormer_2023} and synthetic dataset \cite{ ignatov_replacing_2020} when tested on real handheld smartphone captures.}
    \label{fig:mpinet_combined}
\end{figure}

There has been a rise in smartphone photography, attributed not only to the convenience of having a camera readily available but also to significant advancements in computational imaging technology \cite{delbracio_mobile_2021}. Although smartphone cameras continue to improve, they still face fundamental hardware limitations: smaller sensor sizes, restricted apertures, and compact optics. These constraints lead to reduced spatial resolution, strong noise, color distortions, and various demosaicing artifacts, especially in low-light scenarios where the sensor must operate at higher gains \cite{delbracio_mobile_2021}. Prior work has shown that one way to mitigate many of these shortcomings is through multi-frame (burst) processing strategies, which exploit multiple captures of the same scene to aggregate information that is otherwise lost in a single shot \cite{tsai_multiframe_1984,wronski_handheld_2019}.

At the heart of this work is the problem of multi-frame super-resolution (MFSR), where sub-pixel offsets across a burst of low-resolution (LR) smartphone camera image frames can be leveraged to reconstruct a single, high-resolution (HR) result. The natural hand tremor of a smartphone user, while potentially introducing motion blur or alignment challenges, also provides the very sub-pixel sampling necessary for resolution enhancement~\cite{wronski_handheld_2019}. Although these methods have made notable progress, significant hurdles remain in delivering consistently artifact-free and high-fidelity images, especially when confronted with dynamic scene content or noisy low-light conditions. Small inaccuracies in alignment can cause ghosting artifacts, and a naive fusion of all frames can exacerbate noise if the approach fails to robustly filter out defective or uninformative regions.

A primary obstacle to MFSR lies in obtaining realistic and perfectly aligned LR-HR pairs for algorithm development. Since hardware-based solutions to capture LR and HR bursts simultaneously introduce sensor misalignment, many works resort to synthetic data generation~\cite{bhat_deep_2021-1,ignatov_replacing_2020}, applying random geometric transformations and noise to HR images. However, such synthetic pipelines often fail to reproduce the exact spatio-temporal correlations found in real handheld captures, and purely synthetic noise rarely mimics the complex noise distributions of a real sensor. Consequently, models trained solely on these datasets often struggle to handle the diverse conditions encountered in practice.

We propose an end-to-end MFSR solution that combines a novel synthetic data engine with a base-frame-centric alignment and fusion framework. Our data generation pipeline takes real short-exposure RAW photographs of static scenes acquired at HR and warps them using homographies estimated from independent handheld smartphone captures. The warped images are then downsampled via nearest-neighbor interpolation to preserve sensor-specific noise statistics. Through this approach, we generate LR-HR pairs that retain far more realistic motion patterns and noise characteristics than previous synthetic approaches. In tandem, our multi-scale RAW-to-RGB super-resolution model prioritizes the base frame as the main reference, reducing motion blur and ghosting by aligning other frames relative to this base. The resulting architecture suppresses artifacts when sub-pixel misalignments and scene changes are present, producing sharper images with consistent color fidelity while maintaining the spatial benefits of multiple frames.  

\noindent \textbf{Contributions.} Our primary contributions are summarized as follows. \textbf{(i)} We propose a realistic synthetic data engine for multi-frame super-resolution that better reflects real smartphone handheld captures by preserving sensor-specific noise and temporally correlated camera motion. \textbf{(ii)} We present a novel GAN-based MFSR network that emphasizes a base frame to mitigate ghosting artifacts and robustly integrates information from additional burst frames using channel attention and residual connections, leading to higher resolution and enhanced perceptual quality. Evaluations on both our synthetic test sets and real handheld bursts demonstrate that our method generates sharper, more detailed results compared to existing state-of-the-art MFSR methods~\cite{bhat_deep_2021-1,bhat_mfsr_2021,dudhane_burst_2022,dudhane_burstormer_2023}.

\section{Related Work}
\label{sec:related-work}


\noindent \textbf{Multi-Frame Super-Resolution.}
MFSR leverages information across multiple LR frames of a scene to reconstruct a HR image. Classical methods were applied to DSLR-like cameras and relied on manually derived constraints and closed-form optimizations to fuse these frames under sub-pixel misalignment for image restoration \cite{tsai_multiframe_1984,irani_improving_1991,elad_fast_2001,peleg_improving_1987}. There has also been adjacent work done in video super-resolution that utilizes information across multiple frames \cite{shi_rethinking_2022}. However, estimating motion between multiple frames is often prone to error in real-world settings, especially when dealing with non-global motion, noise, and dynamic scenes.

Recent deep learning methods applied to smartphone camera images have demonstrated notable improvements in MFSR by learning data-driven representations and motion modeling. \cite{wronski_handheld_2019} demonstrated that handheld motion acquired through natural hand tremors provided sub-pixel coverage that can be exploited for super-resolution on a smartphone. \cite{bhat_deep_2021-1} proposed DBSR Network by applying feature alignment and attention-based fusion on synthetic LR-HR bursts. Extending upon \cite{bhat_deep_2021-1}, \cite{bhat_mfsr_2021} introduces additional modules for multi-frame denoising and alignment in the feature space. BIPNet \cite{dudhane_burst_2022} leveraged implicit feature alignment and pseudo-burst generation strategies for handling noisy RAW inputs. More recently, Burstormer \cite{dudhane_burstormer_2023} was proposed with a multi-scale design and flexible inter-frame communication to further enhance MFSR performance. Recently, Vision Transformers \cite{dosovitskiy_image_2020} have been proposed to solve various image restoration tasks, but they are difficult to implement on mobile devices \cite{wang_towards_2022} and ConvNets have been shown to outperform them on smaller datasets \cite{dosovitskiy_image_2020,liu_convnet_2022}.

\vspace{0.4em}
\noindent \textbf{Synthetic Data for Super-Resolution.}
Due to difficulties in collecting aligned LR-HR training pairs in real handheld scenarios, many methods resort to synthetic data generation strategies \cite{bhat_deep_2021-1, ignatov_replacing_2020}. These pipelines typically down-sample HR images and inject artificial noise or use approximate camera models to emulate complex sensor effects. Initial attempts used simple transformations such as uniform random translations and rotations \cite{wronski_handheld_2019, bhat_deep_2021-1}. Nevertheless, synthetic data still exhibits a domain gap when compared to real smartphone captures. Addressing this realism gap in synthetic data generation remains an active area of research.


\section{Synthetic Data Engine}
\label{sec:synthetic_data_generation}
At the time of handheld phone camera capture, natural hand tremors cause each frame of a multi-frame (MF) capture to be slightly offset as frames are sequentially captured. Thus, any handheld MF capture will contain spatial differences between frames owing to differences in how light is captured by the camera's sensor for each frame. Natural hand tremors are highly periodic \cite{marshall_physiological_1956} and generally consistent across the population \cite{sturman_effects_2005}. Recent work has demonstrated that the hand tremor of a user holding a mobile camera is sufficient to provide increased spatial resolution to a sub-pixel accuracy for super-resolution applications \cite{wronski_handheld_2019}.

For super-resolution (SR) applications of MF handheld captures, we require aligned LR-HR captures at a sub-pixel accuracy. However, multi-resolution captures of the same exact scene are not physically possible, such that the only difference between two independent captures is caused by handheld motion. Any attempts to use a LR camera sensor and HR camera-sensor simultaneously to capture an image of the same place at the same time is not possible with sub-pixel accuracy (i.e., there will be sub-pixel offset/misalignment even if two camera sensors are installed side-by-side). In a real-life setting, the only spatial difference between the lower resolution inputs and higher resolution final output should be due to handheld motion of the phone bearer and any object motion, and \textit{not due to misalignment of sensors}. Prior literature has utilized secondary alignment methods to warp misaligned low- and high-resolution images so they are registered \cite{bhat_deep_2021-1}, but this introduces a secondary layer of complexity and interpolation error.

An alternative approach is to synthetically create a MFSR dataset by adding noise and motion. \cite{bhat_deep_2021-1} applied a reverse camera pipeline on a sRGB HR image (i.e., ground truth) before it is down-sampled and added with randomly sampled noise and motion to create a pseudo-handheld LR capture (i.e., input). This process is repeated to create any number of LR burst frames from a single HR image. While this method produces perfectly aligned input-GT pairs, the inputs are derived from the GT and thus only vary due to \textit{purely synthetic} noise and global motion. Camera noise in a real-life setting is highly sensor-specific and can be \textit{temporally and spatially correlated}. In addition, handheld motion is periodic and not fully random. In this paper, we demonstrate that prior noise and motion generation schemes fail in achieving high performance in a real-life handheld multi-frame capture setting and devise a synthetic data generation method based on sensor-specific noise and handheld motion that is more suited to a real-world smartphone multi-frame capture.

\subsection{Static Short-Long Exposure Photography}
To create a synthetic dataset of aligned LR-HR images, we acquired a dataset consisting of registered long exposure and short exposure MF images at HR using a mobile device following \cite{anaya_renoir_nodate, ponomarenko_image_2015}. All images were captured using a tripod to ensure static scenes without any global motion between frames. Utilizing a tripod eliminates unintended camera shake, allowing for perfect alignment between frames.

We fused multi-exposure RAW frames to create our ground truth \cite{anaya_renoir_nodate,ponomarenko_image_2015,xu_multi-exposure_2022}. By using a tripod and fusing multiple frames, we mitigate both noise and blur \cite{hasinoff_burst_2016}. On the short exposure RAW frames, we first create synthetic global motion as described in Section \ref{handheld-motion}. After synthetic motion is added, the short exposure RAW frames are down-sampled 2x using nearest neighbor interpolation to create our LR inputs. We conduct nearest neighbor down-sampling to maintain local noise characteristics inherent in the data capture. Thus, \textit{sensor-specific noise} is preserved in the LR inputs. Through this, we create a dataset of LR-HR pairs which are \textit{perfectly aligned}, contain \textit{real noise characteristic}, and \textit{real simulated handheld motion}, leading to more generalizable and reliable outcomes in our experiments.

\subsection{Global Handheld Motion Modeling} \label{handheld-motion}
\begin{figure}[ht]
  \centering
  \begin{minipage}[b]{0.7\linewidth}
    \centering
    \includegraphics[width=0.75\linewidth]{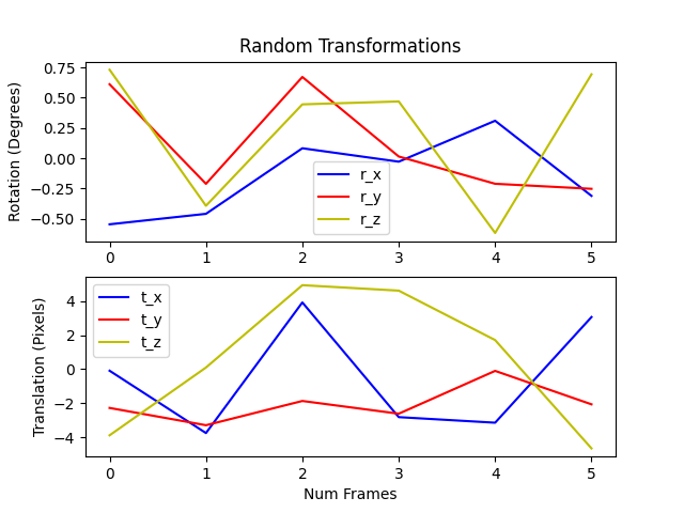}
    \subcaption{Uniform distribution motion}
  \end{minipage}
  \begin{minipage}[b]{0.7\linewidth}
    \centering
    \includegraphics[width=0.75\linewidth]{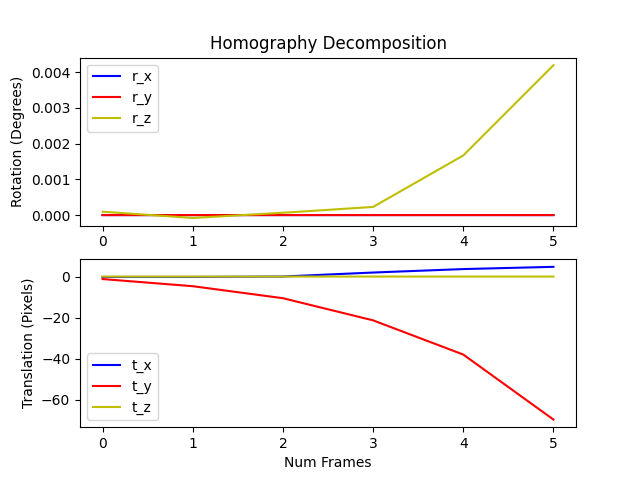}
    \subcaption{Real handheld motion}
  \end{minipage}
  \caption{An example of global motion sampled from a uniform distribution (top) does not contain spatial and temporal correlations found in real handheld motion (bottom). More examples in supplementary materials.}
  \label{fig:handheld_motion_comparison}
\end{figure}

To understand handheld motion when capturing a burst of photographs through a smartphone camera, we examined global motion in a set of 102 MF captures. These bursts were captured under bright-light photography conditions. Prior work has modeled this global motion as uniform translations and rotations \cite{wronski_handheld_2019, bhat_deep_2021-1}. However, we observe with our MF captures, that global motion is temporally and spatially correlated (Figure \ref{fig:handheld_motion_comparison}). Thus, solely sampling displacements from a uniform displacements would not suffice in capturing the true global motion during handheld MF photography. To mimic \textit{real} handheld motion, we sample motion from our MF smartphone captures for photography. We apply the following pipeline to generate global handheld motion that mimics real handheld motion:

\begin{itemize}
    \item Extract all homography matrices \( H_F \) representing the perspective warp to transform a base frame (i.e., first frame) into a handheld burst frame \( F_i \) for all given multi-frame captures in an independent handheld data capture. All elements of \( H_F \) describe the transformation to go from a base frame to a non-base frame.  Thus, if \( (x_1, y_1) \) are coordinates of a base frame \( F_1 \), and \( h_{j,k} \) represent the affine transforms of \( H_F \), then the coordinates \( (x_i, y_i) \) of a reference frame \( F_i \) can be calculated using the following perspective warp:
\begin{equation}
        \begin{bmatrix} x_i \\ y_i \\ 1 \end{bmatrix} = H_F \begin{bmatrix} x_1 \\ y_1 \\ 1 \end{bmatrix}, \quad H_F = \begin{bmatrix} h_{11} & h_{12} & h_{13} \\ h_{21} & h_{22} & h_{23} \\ h_{31} & h_{32} & 1 \end{bmatrix} 
    \end{equation}
    \item For each IN-GT patch, randomly sample the homography matrices \( H_F \) and apply a perspective warp on the IN frame along the \textit{center} of the frame. The base frame (\textit{F} = 0) is not warped, but all subsequent frames are warped by sequentially sampling \( H_F \).

\end{itemize}

Thus, from a dataset of real handheld captures, we sample homography matrices \( H_F \) representing the transformation from base frame to frame \textit{F} (i.e., real handheld motion), and use it to warp a sequence of short-exposure static frames, thereby inducing synthetic global motion that has the same characteristics of a handheld smartphone MF capture. Since the matrices are sampled independently, temporal correlations are not captured in this model because it would require extensive data collection in order to generate independent groups of homographies. We instead rely on modelling the marginal probability distributions of homographies of true handheld motion, but we note the importance of joint distributions for future work.

\section{MFSR-GAN}
\label{sec:mfsr-net}
\begin{figure*}
    \centering
        \includegraphics[width=0.64\linewidth]{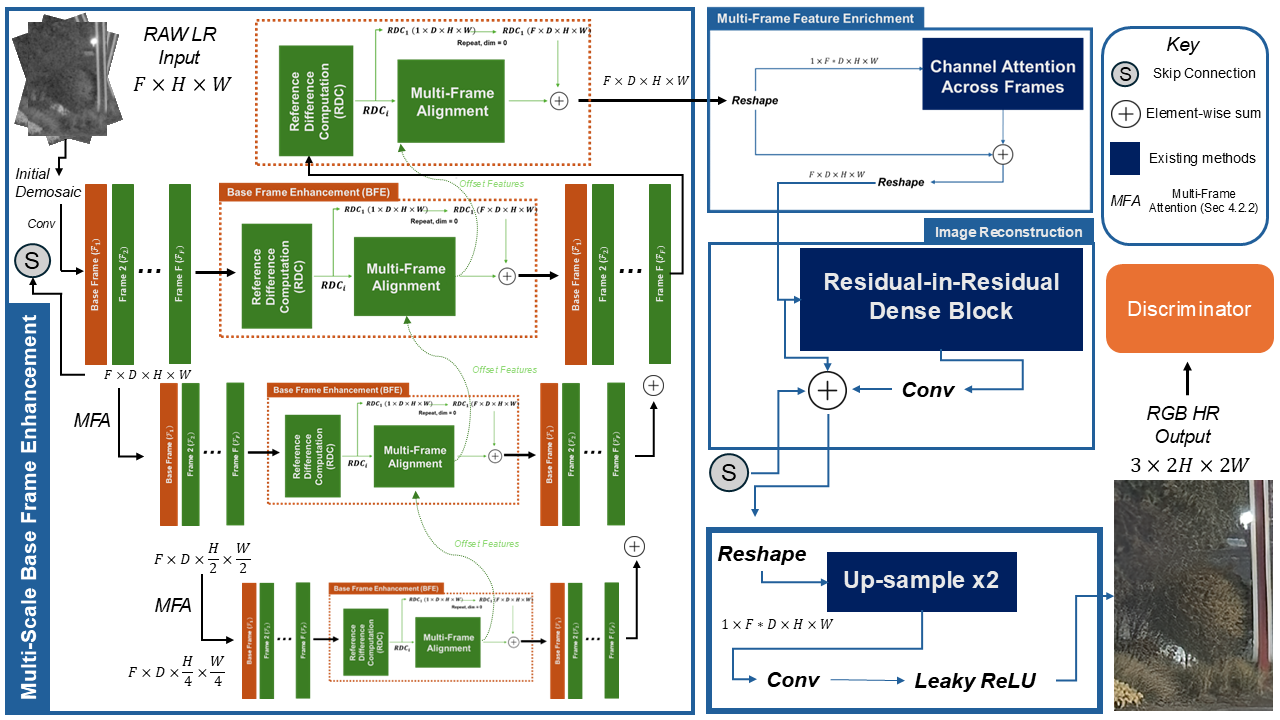}
        
    
    \caption{Overview of MFSR-GAN. Descriptions of modules in Section \ref{sec:mfsr-net} and further details in supplementary materials.}
    \label{fig:mpinet_combined}
\end{figure*}

We develop MFSR-GAN which inputs multiple RAW noisy image frames at LR captured via a handheld smartphone, and outputs a single RGB image at HR. Our model initializes a nearest neighbor demosaic of RAW inputs (Section \ref{demosaic}), aligns frames together while emphasizing the base frame (Section \ref{bfe}), and fuses frames together (Section \ref{mff}) before reconstructing and up-sampling the final output image (Section \ref{reconstruction}). An overview of our generator architecture is shown in Figure \ref{fig:mpinet_combined}. The design of our generator network expands upon prior work on MFSR \cite{dudhane_burstormer_2023, bhat_deep_2021-1}. GANs have been adapted for many image synthesis applications, from natural images \cite{brock_large_2019} to microscopes \cite{chen_single_2024}. Several state-of-the-art techniques for single-image super-resolution also utilize GANs \cite{yang_deep_2019} but GANs have not previously been significantly developed for MFSR.


\subsection{Demosaicking Initialization} \label{demosaic}
The network takes as input a multi-frame sequence $\{\mathbf{I}_i\}_{i=1}^{F}$ of $F$ RAW image frames, each of size $H \times W$. We initially demosaic our RAW input to RGB via a simple nearest neighbor demosaic. If $(x, y)$ denote pixel coordinates, through demosaicking, we reconstruct the noisy RGB image $I_{\text{RGB}}(x, y) = [R(x, y), G(x, y), B(x, y)]^T$ where $\mathbf{I}_i^{\text{RGB}} \in \mathbb{R}^{{H} \times {W} \times 3}$. Initially, we interpolate the green ($G$) pixel values since $G$ pixels are sampled at twice the density of red ($R$) and blue ($B$) pixels. Next, we compute the color differences ($R - G$ and $B - G$) at the locations of the $R$ and $B$ pixels and perform interpolation on these differences. Following this, we add the interpolated green image to the interpolated color difference image to obtain the final $R$ and $B$ images. This demosaicking approach leverages the spatial proximity of similar color pixels to fill in missing color information, providing a better than random initialization of the full-color image before further processing by the network. We argue that an explicit demosaic at this early stage allows the model to emphasize spatial differences between LR inputs and HR labels, and thus be more effective at downstream alignment and fusion tasks. Following an initial demosaic, we map each multi-frame input $\{\mathbf{I}_i\}_{i=1}^{F}$ to a deep feature representation by passing through a convolutional layer, achieving an encoding $\{\mathbf{I}_i\}_{i=1}^{F} \in \mathbb{R} ^{W \times H \times D}$ where $D=64$ for our experiments.

\subsection{Multi-Scale Base Frame Enhancement} \label{bfe}
The multi-scale base frame enhancement module aims to align and enhance the base (reference) frame using information from the non-reference frames across multiple scales. A multi-scale approach to image restoration was first proposed by \cite{zamir_restormer_2022}, and applied in a MFSR application by \cite{dudhane_burstormer_2023}. Similar to \cite{dudhane_burstormer_2023}, we operate at three scales $s \in \{0,1,2\}$, where each scale corresponds to a different spatial resolution.

However, we design our alignment module to emphasize base frames and overcome motion blur artifacts (or "ghosting"). To do so, we apply reference difference computational and multi-frame alignment recursively. At each scale \textit{s}, we explicitly emphasize our base frame during alignment.

\subsubsection{Reference Difference Computation (RDC)}
We designate the first frame $\mathbf{I}_1$ as the base frame. We compute the feature (\(\mathfrak{F}\)) differences between each non-reference frame and the base frame (\textit{i} = 1) for each scale level \textit{s}:

\begin{equation}
\mathbf{RDC}_i^{(s)} = 
\begin{cases} 
\mathbf{\mathfrak{F}}_1^{(s)}, & \text{if } i = 1 \\
\mathbf{\mathfrak{F}}_i^{(s)} - \mathbf{\mathfrak{F}}_{1}^{(s)} & \text{if } i = 2, \dots F
\end{cases}
\end{equation}

The residuals $\mathbf{RDC}_i^{(s)}$ highlight misalignments and motion between frames, forcing the network to prioritize the base frame in absolute terms and thus only learn spatial differences\textit{ relative} to the base frame.

\subsubsection{Multi-Frame Alignment \& Attention}
The Multi-Frame Alignment module aligns non-reference frames to the base frame using deformable convolutions and attention mechanisms. We employ deformable convolutions~\cite{dai_deformable_2017} to handle complex motions:

\begin{equation}
\tilde{\mathbf{\mathfrak{F}}}_i^{(s)}(\mathbf{x}) = \sum_{\mathbf{\delta} \in \mathcal{N}} \mathbf{w}(\mathbf{\delta}) \cdot \mathbf{RDC}_i^{(s)}(\mathbf{x} + \mathbf{\delta} + \Delta \mathbf{p}_i^{(s)}(\mathbf{x}, \mathbf{\delta})),
\end{equation}

where $\mathcal{N}$ represents the convolution kernel neighborhood, $\mathbf{w}(\mathbf{\delta})$ are the learned kernel weights, and $\Delta \mathbf{p}_i^{(s)}$ are offset predictions:  

\begin{equation}
\Delta \mathbf{p}_i^{(s)} = \text{Conv}_{\text{offset}}(\mathbf{RDC}_i^{(s)}),
\end{equation}

with $\mathbf{RDC}_i^{(s)}$ now also incorporating features from the current scale as well as the offset features propagated from the previous scale. This integration allows the module to refine offset predictions progressively across scales, improving alignment precision for complex motion patterns.

To refine alignment, we expand upon the multi-frame attention mechanism introduced in \cite{dudhane_burstormer_2023} and propose Multi-Frame Attention (Figure 3 in supplementary materials) to further improve computational efficiency. As depicted in Figure \ref{fig:mpinet_combined}, we utilize a simple gate as described by \cite{avidan_simple_2022} followed by residual channel attention as described by \cite{zhang_rcan}. This halves the feature dimensionality. We keep the gated d-conv feedforward network in \cite{dudhane_burstormer_2023} (originally described by \cite{zamir_restormer_2022}) to perform controlled feature transformation by emphasizing more important features to be passed through to the next module.

We also incorporate skip connections to preserve original information as depicted in Figure \ref{fig:mpinet_combined} . In addition, we introduce a \textit{ghosting} skip connection, whereby we add a skip from \textit{only} the base frame feature ($i = 1$) to after Multi-Frame Alignment:

\begin{equation}
\hat{\mathbf{\mathfrak{{F}}}}_i^{(s)} = \mathbf{\mathfrak{\tilde{F}}}_i^{(s)} + \mathbf{\mathfrak{F}}_1^{(s)}, \quad \text{for } i = 2, \dots, F
\end{equation}

Combining deformable convolutions with simple attention mechanisms enables effective modeling of complex motions while focusing on relevant features, enhancing alignment quality. Skip connections with respect to base frame also allow for minimizing ghosting artifacts.

\subsection{Multi-Frame Feature Enrichment}
\label{mff}

After alignment, we merge information from all frames to construct a spatially-enriched feature representation combining all frames. This module extends prior work which has found success in single frame super-resolution tasks by using long range skip connections with channel attention \cite{zhang_rcan}. At Scale 0, we concatenate aligned features $\{\mathfrak{\hat{F}}_i^{(0)}\}_{i=1}^{F} \in \mathbb{R} ^{W \times H \times D}$  along the channel dimension to create a new feature representation \(\mathfrak{\hat{F}}_{concat}^{(0)}\in \mathbb{R} ^{W \times H \times D*F}\). We treat this representation as a ``single image'' and apply channel attention \cite{zhang_rcan} across all features of all frames. We also add a long skip connection. Different frames may provide spatio-temporal advantages and the combined spatial resolution of multiple frames will always be $\geq$ a single frame. Thus, multi-frame feature attention allows the model to access long-range context, thereby learning important spatially-enriched features across \textit{and} within frames.

\subsection{Image Reconstruction}
\label{reconstruction}
Once cross-frame features are attended to, our final aim is to maximize the overall perceptual quality. Our reconstruction module is designed to effectively capture both low-level and high-level features, facilitating the generation of a final HR image with rich details and textures.

The first step is to reconstruct individual frames separately. We adapt the residual-in-residual dense blocks (RRDB) from \cite{leal-taixe_esrgan_2019} for our image reconstruction. Each RRDB consists of a series of convolutional layers where each layer takes as input the concatenation of all preceding layers within the block. We stack three such layers connected via residual connections, enabling hierarchical learning of features. Given a single input frame $\mathfrak{\hat{F}}_0^{(0)}\ \in \mathbb{R} ^{W \times H \times D}$, the output of the $n$-th convolutional layer within the RRDB is computed as:

\begin{equation}
\tilde{\mathfrak{R}}_{n}^{(0)} = \sigma\left( \mathbf{W}_n * \left[\mathfrak{\hat{F}}_{0}^{(0)}, \mathfrak{\hat{F}}_{1}^{(0)}, \dots, \mathfrak{\hat{F}}_{n-1}^{(0)} \right] \right) + \mathfrak{\hat{F}}_0^{(0)}
\end{equation}

where $\mathbf{W}_n$ are weights of the $n$-th convolutional layer, $*$ denotes the convolution operation, $[\cdot]$ indicates channel-wise concatenation, and $\sigma(\cdot)$ is the Leaky ReLU activation function with a negative slope $\alpha=0.2$. From this, a single block of our RRDB implementation becomes

\begin{equation}
\mathfrak{R}_{\text{RRDB}}^{(0)} = \gamma \cdot \tilde{\mathfrak{R}}_{n,3}^{(0)}\left( \tilde{\mathfrak{R}}_{n,2}^{(0)}\left( \tilde{\mathfrak{R}}_{n,1}^{(0)}\left(\mathfrak{R}_0^{(0)} \right) \right) \right) + \mathfrak{R}_0^{(0)}
\end{equation}

where $\tilde{\mathfrak{R}}_{n,k}^{(0)}(\cdot)$ denotes the function of the $k$-th $n$-layered residual dense block \cite{leal-taixe_esrgan_2019}, and $\gamma=0.2$ is a scaling factor. This multi-level residual learning strategy allows the network to capture a wide range of features and eases the flow of gradients during training.

Through our RRDB layers, we reconstruct individual frames. Following this, we fuse them into a single frame while up-sampling to to the desired HR size. Based on \cite{leal-taixe_esrgan_2019}, such a RRDB setup contains sufficiently rich embeddings and thus, we opt for a simple up-sampling and fusion strategy by using a combination of nearest-neighbor interpolation and convolutional operations. This helps reduce our total compute and network parameters, as opposed to pixel shuffle \cite{pixelshuffle_2016} previously used in \cite{dudhane_burstormer_2023} which is more computationally costly.

\subsection{Discriminator}\label{reconstruction}
We utilize a relativistic discriminator \cite{jolicoeur-martineau_relativistic_2018} as implemented in \cite{leal-taixe_esrgan_2019}. In summary, if a standard discriminator is $D(x) = \sigma(C(x))$, a relativistic discriminator ($D_{R}$) will be
\[
D_{R}(x_r, x_f) = \sigma\big(C(x_r) - \mathbb{E}_{x_f}[C(x_f)]\big),
\]
where $\mathbb{E}_{x_f}$ averages over fake samples. The relativistic losses for the discriminator ($D$) and generator ($G$) thus are
\begin{equation}
L_{D}^{R} = -\mathbb{E}_{x_r}[\log D_{R}] - \mathbb{E}_{x_f}[\log(1 - D_{R})],
\end{equation}
\begin{equation}
L_{G}^{R} = -\mathbb{E}_{x_r}[\log(1 - D_{R}] - \mathbb{E}_{x_f}[\log D_{R}]
\end{equation}
where $x_f = G(x_i)$ for input LR image $x_i$ and $x_r = G(y_i)$ for HR labels $y_i$, thereby enabling the generator to use gradients from both real and fake data, enhancing edge sharpness and texture detail.

\section{Experiments}
\label{sec:experiments}

\begin{figure*}[t]
    \centering
    \includegraphics[width=0.994\linewidth]{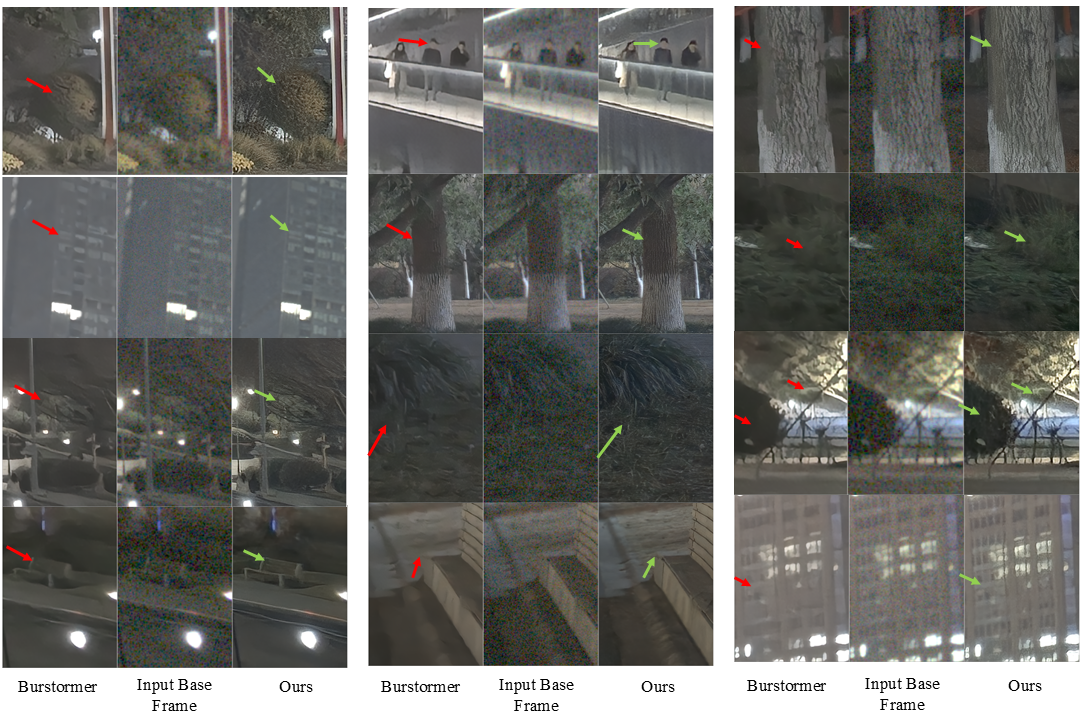}
    \caption{Multi-frame super-resolution qualitative results for real handheld burst photography under low-light conditions for MFSR-GAN and state-of-the-art Burstormer \cite{dudhane_burstormer_2023}. Both models fully trained using our synthetic data engine.}
    \label{fig:visual_results1}            
\end{figure*}

\begin{figure}
  \centering
  \includegraphics[width=\columnwidth]{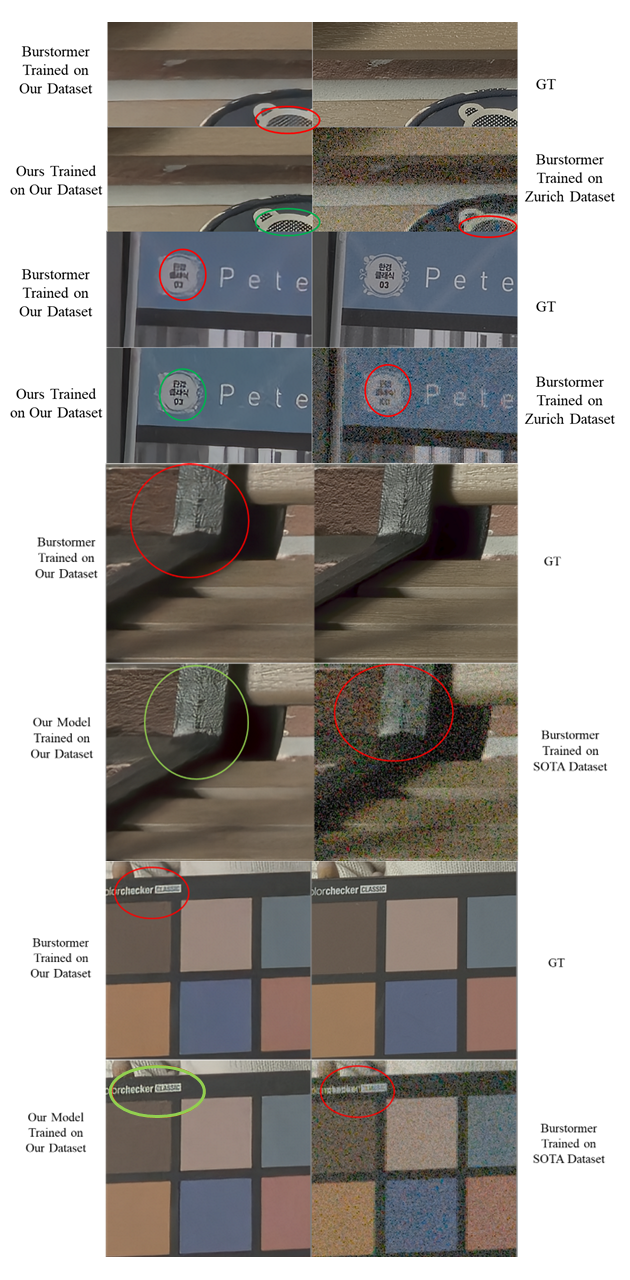}
  \caption{Qualitative comparison of our model and synthetic dataset against SOTA synthetic dataset \cite{ignatov_replacing_2020} and model \cite{dudhane_burstormer_2023}.}
  \label{fig:qualitative_results}
\end{figure}

\begin{figure}
  \centering
  \includegraphics[width=\columnwidth]{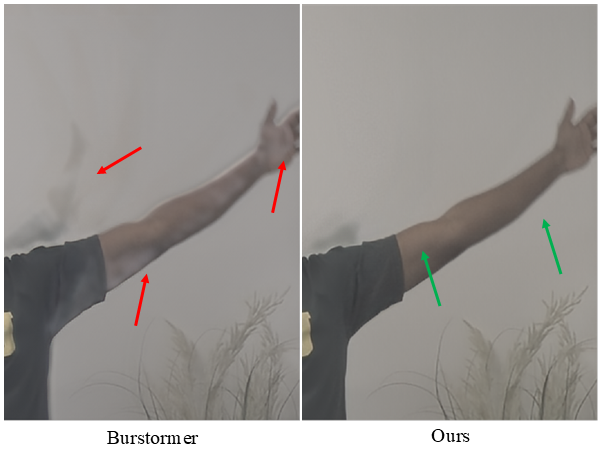}
  \caption{Qualitative comparison of motion handling for our model against Burstormer \cite{dudhane_burstormer_2023} on real handheld MF capture. Person is waving their hands throughout MF capture. Both models trained on our synthetic dataset.}
  \label{fig:model_qualitative_results}
\end{figure}

\begin{table*}
  \centering
  \begin{tabular}{@{}lccc|ccc@{}}
    \toprule
    \textbf{Methods} & \multicolumn{3}{c}{\textbf{Our Synthetic Dataset}} & \multicolumn{3}{c}{\textbf{Zurich Synthetic Dataset \cite{ignatov_replacing_2020}}} \\
    \cmidrule(lr){2-4} \cmidrule(lr){5-7}
    & \textbf{SSIM $\uparrow$} & \textbf{LPIPS $\downarrow$} & \textbf{PSNR $\uparrow$} & \textbf{SSIM $\uparrow$} & \textbf{LPIPS $\downarrow$} & \textbf{PSNR $\uparrow$} \\
    \midrule
    Burstormer \cite{dudhane_burstormer_2023} & 0.884 & 0.364 & 31.1 & \textbf{0.994} & \textbf{0.027} & \textbf{41.3} \\
    DBSR \cite{bhat_deep_2021-1} & 0.883 & 0.361 & \textbf{32.9} & - & - & - \\
    MFIR \cite{bhat_mfsr_2021} & 0.882 & 0.354 & \textbf{32.9} & - & - & - \\
    \midrule
    \textbf{Ours} & \textbf{0.896} & \textbf{0.321} & 31.7 & \textbf{0.994} & 0.037 & 40.9 \\
    \bottomrule
  \end{tabular}
  \caption{Quantitative comparison across models for Zurich synthetic dataset \cite{ignatov_replacing_2020} and our synthetic dataset.}
  \label{tab:comparison}
\end{table*}

We evaluate the performance of our proposed \textbf{a)} synthetic data generation framework, and \textbf{b)} MFSR-GAN on end-to-end multi-frame super-resolution. Additional experiments with alternative design choices and training settings, along with more visual results and model details are provided in the supplementary materials section.

\label{training}\textbf{Setup.} We separately train MFSR models (Table \ref{tab:comparison}) for end-to-end multi-frame super-resolution without pre-training on both our proposed synthetic data engine as well as the existing MFSR Zurich RAW-to-RGB synthetic dataset \cite{ignatov_replacing_2020, bhat_deep_2021-1}. We keep the same overall implementation and training setups for each model as originally described, but adjust the inputs to accept 8 input frames of 4-channel RGGB RAW images to be compatible with our proposed data generation framework. We also train MFSR-GAN on both the Zurich dataset as well as our proposed dataset. We train our generator model with weighted L1 loss, MS-SSIM loss \cite{wang_multiscale_2003}, and relativistic generator loss (Section \ref{sec:mfsr-net}). We train our discriminator with a relativistic discriminator loss (Section \ref{sec:mfsr-net}). We use Adam optimizer with an $5 \times 10^{-4}$ initial learning rate that is reduced to $5 \times 10^{-6}$ through a cosine annealing schedule \cite{loshchilov_sgdr_2017} with a batch size of four. Training was carried out on four V100 GPUs for 100 epochs.

\label{synthetic-dataset-experiments}\textbf{Synthetic Dataset.} The Zurich RAW to RGB dataset was first released by \cite{ignatov_replacing_2020} and used to create synthetic ``bursts" for MFSR by \cite{bhat_deep_2021-1}. Synthetic frames were created by applying random bilinear translations and rotations to linear RGB images after they were converted to raw sensor values. After this, the images were downsampled by another bilinear kernel to obtain the multi-frame LR RGB inputs. Following this, read and shot was added before two color channels per pixel were discarded to obtain the Bayer CFA mosaicked RAW bursts.

As described in Section \ref{sec:synthetic_data_generation}, we capture independent RAW short-exposure images at HR that are warped in the RGB space using a homography of real handheld multi-frame smartphone captures, thereby ensuring real spatial and temporal correlations between frames are preserved. Next, input frames are downsampled using nearest neighbor interpolation to ensure local (real) noise characteristics are also preserved. Thus, we do not add synthetic noise as our inputs already contain representative noise characteristics of a real LR input. We create two variations of our synthetic dataset: low-light and bright-light. For our low-light version, we capture 376 short-long exposure pairs, while for our bright-light dataset, we capture 107 short-long exposure pairs. All images are static, captured on a tripod using a smartphone at 50 MP (HR). The corresponding LR inputs are 12 MP. We train on 256x256 image patches. For validation, we captured thirty handheld 12 MP RAW images using the same smartphone. Validation images were captured for the purposes of photography and thus contain all distortions, artifacts and motion as would be expected in a real handheld smartphone capture. Since these images represent ``real" handheld photography, one-to-one real ground truth images cannot be captured and qualitative evaluation is performed as shown in Figures \ref{fig:visual_results1} and \ref{fig:model_qualitative_results}.

Table \ref{tab:comparison} demonstrates that models perform relatively competitively on the Zurich synthetic dataset \cite{ignatov_replacing_2020}, but the overall performance is poorer for our synthetic test set. We argue this is because our synthetic dataset contains more realistic perturbations and noise, and is thus a much harder task to excel in. We have added more results in supplementary material showing the importance of accurate motion modeling. Figure \ref{fig:qualitative_results} shows qualitative results of the state-of-the-art MFSR model Burstormer \cite{dudhane_burstormer_2023} achieving better noise handling on our dataset as compared to the Zurich dataset \cite{ignatov_replacing_2020} when trained identically.

\label{synthetic-dataset-experiments}\textbf{MFSR-GAN Network.} We train our model along with DBSR \cite{bhat_deep_2021-1}, MFIR \cite{bhat_mfsr_2021} and Burstormer \cite{dudhane_burstormer_2023} as described in their original literature with the training settings described in Section \ref{training}. Table \ref{tab:comparison} shows that we outperform other models in perceptual image quality on our synthetic dataset, while demonstrating non-inferior competitive performance on the Zurich dataset. We emphasize the base frame during our frame fusion strategy as compared to prior literature. We present visual comparisons on a real handheld multi-frame capture in Figures \ref{fig:visual_results1} and \ref{fig:model_qualitative_results}. We show that images generated by our approach are sharper, contain higher details, and significantly lower motion blur artifacts, a limitation of prior methods.

\section{Conclusion}\label{sec:conclusion}
We tackle the challenge of real-world multi-frame super-resolution for smartphone photography by jointly addressing the difficulties of realistic synthetic data generation and robust frame fusion. To this end, we introduce a novel synthetic data engine designed to preserve genuine sensor-specific noise patterns and capture the natural spatio-temporal correlations arising from handheld camera motion. This new pipeline goes beyond conventional random transformations and noise injection, thus reducing the realism gap between synthetic training data and real smartphone bursts. We also propose MFSR-GAN for RAW-to-RGB multi-frame super-resolution. Quantitative evaluations on both our own synthetic dataset and established benchmarks demonstrates competitive performance. Experiments on real handheld bursts consistently shows that MFSR-GAN balances noise suppression, resolution enhancement, and artifact mitigation more effectively than existing methods.

\setcounter{page}{1}
\maketitlesupplementary

\section{Synthetic Data Engine}
\begin{figure*}[!b]
  \centering
    \includegraphics[width=\linewidth]{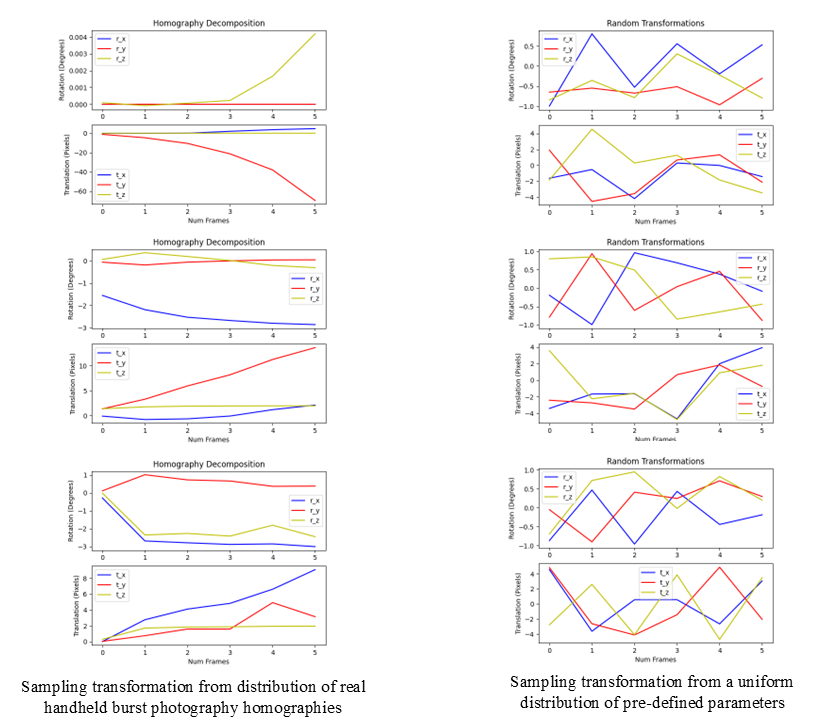}
  \caption{Sampling transformations from a uniform distribution of pre-defined rotations and translation parameters (right) versus sampling transformations from a dataset of homography matrices defining the transformations within a real handheld burst photography capture (left). Uniform distribution sampling does not contain spatial and temporal correlations found in real handheld motion.}
  \label{fig:handheld_motion_comparison1}
\end{figure*}

\begin{figure*}[h]
  \centering
    \includegraphics[width=0.8\linewidth]{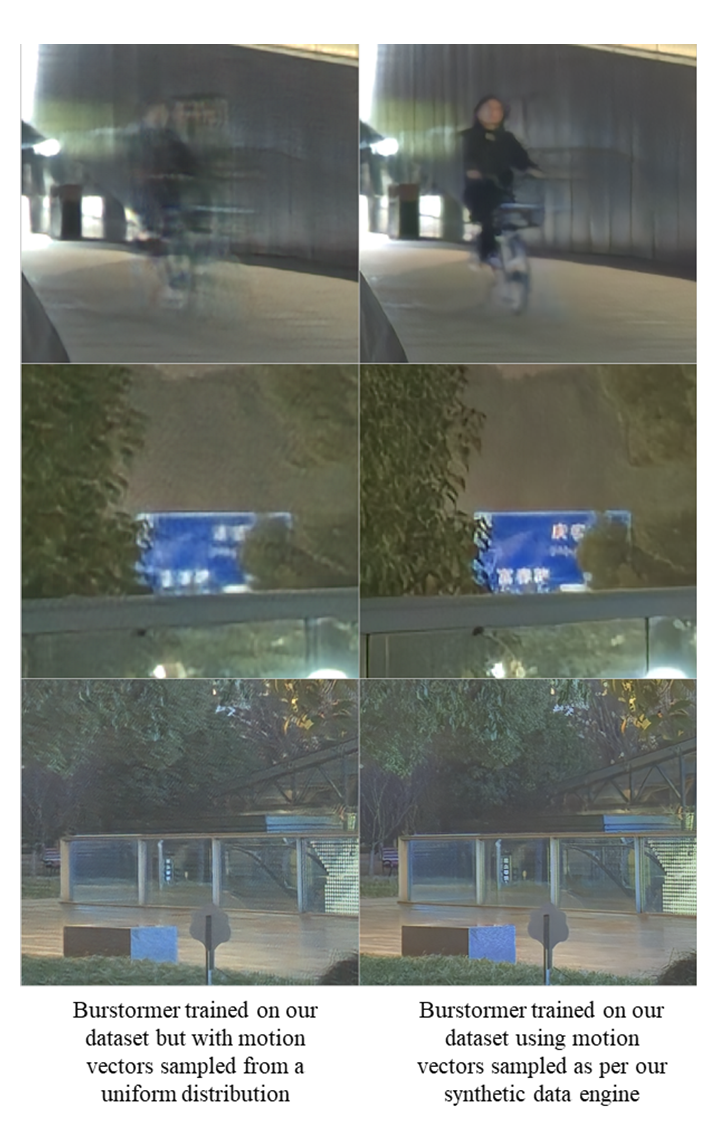}
  \caption{Burstormer \cite{dudhane_burstormer_2023} trained on our dataset with motion vectors sampled from a uniform distribution of pre-defined rotations and translation parameters (left) versus Burstormer trained on the same dataset but with motion vectors sampled from a dataset of homography matrices defining the transformations within a real handheld burst photography capture as per our synthetic data engine (right). All experimental conditions identical.}
  \label{fig:ablation_random_vs_ours}
\end{figure*}

We provides more details of our synthetic data egnine. Figure \ref{fig:handheld_motion_comparison1} provides more examples of the temporal and spatial correlation of motion within a multi-frame data capture. Figure \ref{fig:ablation_random_vs_ours} shows visual results for the SOTA model in MFSR \cite{dudhane_burst_2022} trained on our dataset with motion sampled from a uniform distribution of rotations and translations against sampling from a dataset of real handheld homography matrices.

\clearpage

\section{Model Architecture Details}
\begin{figure*}[t]
    \centering
    \begin{subfigure}[b]{0.49\textwidth}
        \centering
        \includegraphics[height=6cm, keepaspectratio]{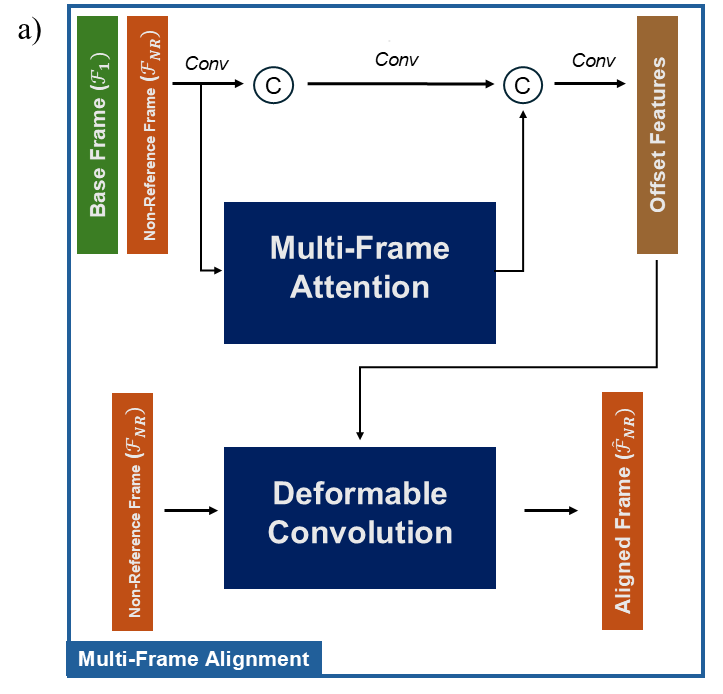}
        \label{fig:lasttwo_archs}
    \end{subfigure}
    \begin{subfigure}[b]{0.49\textwidth}
        \centering
        \includegraphics[height=6cm, keepaspectratio]{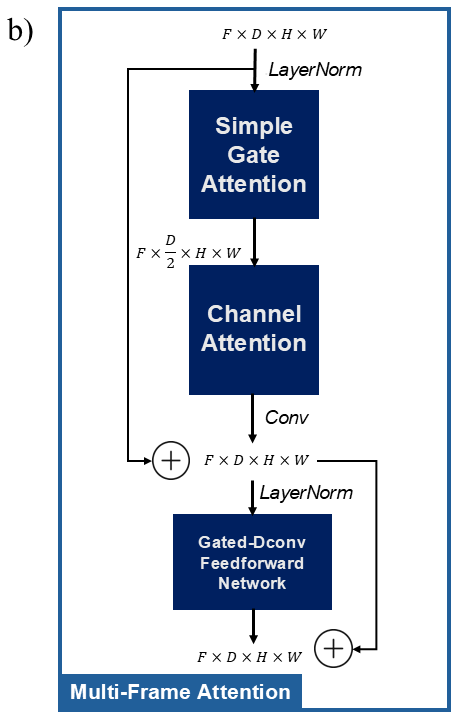}
        \label{fig:firsttwo_archs}
    \end{subfigure}
    
    \caption{MFSR-GAN modules for (a) Multi-Frame Alignment, and (b) Multi-Frame Attention. Descriptions in Section \ref{sec:mfsr-net}.}
    \label{fig:additional_mfsrgan_modules}
\end{figure*}

We provide more details for our model architecture previously introduced in Section 4. Our model architecture builds upon the work previously completed for MFSR by \cite{dudhane_burstormer_2023}. Our adaptions of the Multi-Frame Alignment and Multi-Frame Attention modules are presented in Figure \ref{fig:additional_mfsrgan_modules} with descriptions in Section 4.2.2.







\clearpage
{
    \small
    \bibliographystyle{ieeenat_fullname}
    \bibliography{references}
}

\end{document}